\relax
\documentclass[letterpaper]{article} 
\usepackage{aaai22}  
\usepackage{times}  
\usepackage{helvet}  
\usepackage{courier}  
\usepackage[hyphens]{url}  
\usepackage{graphicx} 
\usepackage{natbib}  
\usepackage{caption} 
\DeclareCaptionStyle{ruled}{labelfont=normalfont,labelsep=colon,strut=off} 
\usepackage{algorithm}
\usepackage{algorithmic}
\usepackage{subfig}
\usepackage{amsmath}

\usepackage[utf8]{inputenc}

\usepackage{xcolor}
\usepackage{booktabs}
\usepackage{inconsolata}
\usepackage{tikz}
\usepackage{pgfplots}
\usepackage{microtype}
\usepackage{amssymb}
\usepackage[titletoc]{appendix}

\newcommand{\name}{\textsc{NumSynth}}
\newcommand{\magicpopper}{\textsc{MagicPopper}}
\newcommand{\popper}{\textsc{Popper}}
\newcommand{\popperplus}{\textsc{Popper+}}

\newcommand{\ale}{\textsc{Aleph}}

\newtheorem{definition}{Definition}
\newtheorem{example}{Example}

\usepackage{pgfkeys}
    \newenvironment{customlegend}[1][]{%
        \begingroup
        \csname pgfplots@init@cleared@structures\endcsname
        \pgfplotsset{#1}%
    }{%
        \csname pgfplots@createlegend\endcsname
        \endgroup
    }%
    \def\addlegendimage{\scriptsize\csname pgfplots@addlegendimage\endcsname}
\usepackage{comment}
\usepackage{newfloat}
\usepackage{listings}
\floatstyle{ruled}
\newfloat{listing}{tb}{lst}{}
\floatname{listing}{Listing}
\title{Relational program synthesis with numerical reasoning}
\author{
    C\'{e}line Hocquette,
    Andrew Cropper
}

\affiliations {
    University of Oxford\\
    celine.hocquette@cs.ox.ac.uk, andrew.cropper@cs.ox.ac.uk
}
\usepackage{bibentry}

\newtheorem{proposition}{Proposition}
\newtheorem{proof}{Proof}

\usepackage{xcolor}

\definecolor{pixel 0}{HTML}{FFFFFF}

\begin{document}

\maketitle

\begin{abstract}
Program synthesis approaches struggle to learn programs with numerical values.
An especially difficult problem is learning continuous values over multiple examples, such as intervals.
To overcome this limitation, we introduce an inductive logic programming approach which combines relational learning with numerical reasoning. 
Our approach, which we call \name{}, uses satisfiability modulo theories solvers to efficiently learn programs with numerical values. Our approach can identify numerical values in linear arithmetic fragments, such as real difference logic, and from infinite domains, such as real numbers or integers. 
Our experiments on four diverse domains, including game playing and program synthesis, show that our approach can (i) learn programs with numerical values from linear arithmetical reasoning, and (ii) outperform existing approaches in terms of predictive accuracies and learning times.
\end{abstract}

\section{Introduction}

\begin{figure}[t]
\small
\centering
\subfloat[Examples]{\label{fig:examples}
\small
\begin{tabular}{l|l}
Positive example & Negative example\\
 \includegraphics[height=70pt]{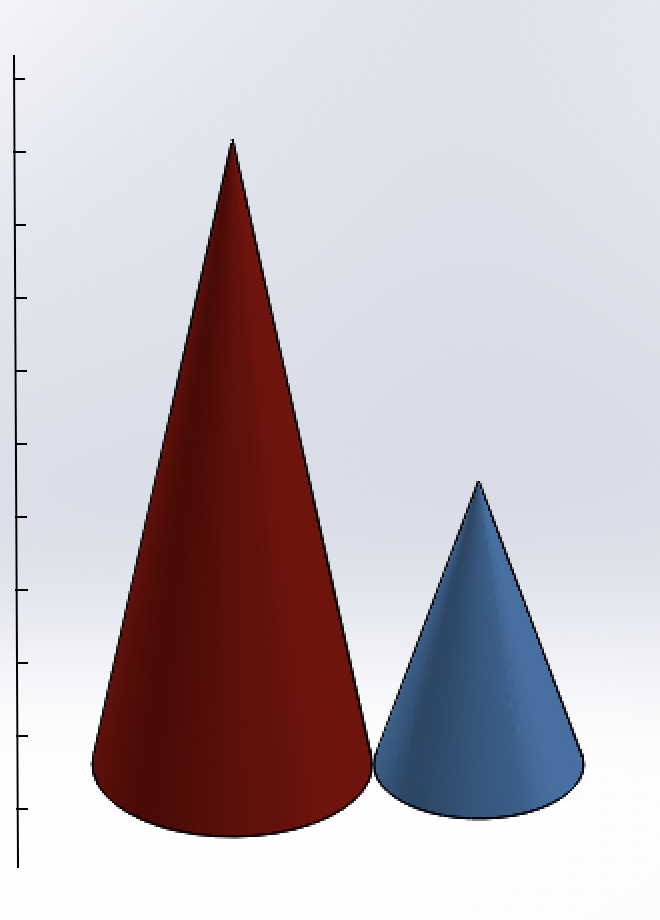} & \includegraphics[height=70pt]{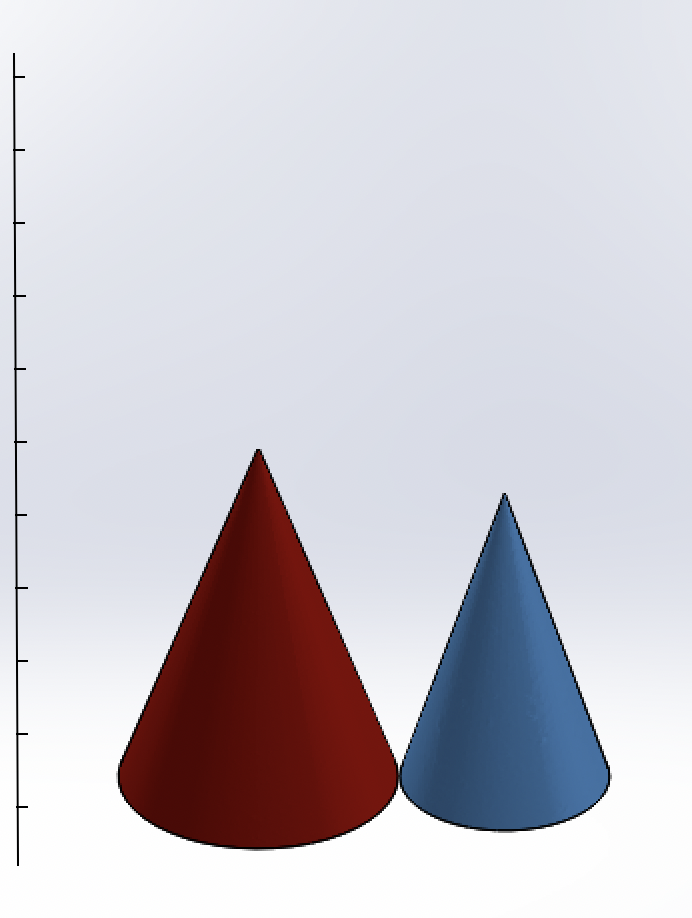} \\ 
\end{tabular}
}\\
\subfloat[A rule which states that two pieces are in contact, one with size \emph{greater or equal} than 7.]{\label{fig:target}
\small
\begin{tabular}{l}
\texttt{zendo(A) $\leftarrow$ piece(A,B), contact(B,C),}\\
\hspace{51pt} \texttt{size(C,D), \textbf{geq}(D,\textbf{7}).}\\
\end{tabular}
}\\
\subfloat[Intermediate hypothesis.]{\label{fig:intermediate}
\begin{tabular}{l}
\texttt{zendo(A) $\leftarrow$ piece(A,B), contact(B,C), size(C,D),}\\
\hspace{51pt} \texttt{\textbf{geq}(D,N), @numerical(N).}\\
\end{tabular}
}
    \caption{Learning a hypothesis for Zendo. Learning this hypothesis involves reasoning with the numerical predicate \emph{geq} represented in bold to identify the numerical value \emph{7}.
    }
\end{figure}

Zendo is a game in which one player, the Master, creates a rule for structures that the rest of the players, as Students, try to discover by building and studying structures which follow or break the rule.
The first student to correctly guess the rule wins.
For instance, suppose the structure on the left of Figure \ref{fig:examples} follows the secret rule while the one on the right does not. 
Figure \ref{fig:target} shows a possible secret rule.
It states that structures must have two pieces in contact, one with size at least 7. 
Discovering this rule involves identifying the numerical value \emph{7}.

Suppose we want to use machine learning to play Zendo, i.e. to learn secret rules from examples of structures.
Then we need an approach that can (i) learn explainable rules, and (ii) generalise from small numbers of examples. However, these requirements are difficult for standard machine learning techniques, yet are crucial for many real-world problems \cite{ilp30}.

Inductive logic programming (ILP) \cite{muggleton1991} is a form of machine learning that can learn explainable rules from small numbers of examples. 
Existing ILP techniques could, for instance, learn rules for simple Zendo problems.
However, existing approaches struggle to learn rules that require identifying numerical values from infinite domains \cite{corapi2011,evans2018,cropper2020b}.
Moreover, although some ILP approaches can learn programs with numerical values \cite{muggleton1995,hocquette2022}, they cannot perform complex numerical reasoning, such as identifying numerical values by reasoning over multiple examples jointly.
For instance, they struggle to learn that the size of one piece must be greater than some particular numerical value, or that the sum of the coordinates describing the position of a piece must be lower than some particular numerical value.
These limitations are not specific to ILP and, as far as we are aware, apply to all current program synthesis approaches \cite{raghothaman2019,ellis2018,crossbeam}.

To overcome these limitations, we introduce an approach that can identify numerical values from infinite domains and reason from multiple examples. 
The key idea of our approach is to decompose the learning task into two stages (i) \emph{program search}, and (ii) \emph{numerical search}. 

In the \emph{program search} stage, the learner searches for partial hypotheses (sets of rules) with variables in place of numerical symbols. 
This step follows \ale's lazy evaluation procedure \cite{srinivasan1999}.
For example, to learn a rule for Zendo, the learner may generate the partial hypothesis shown in Figure \ref{fig:intermediate}. In this hypothesis, the first-order variable $N$ is marked as a numerical variable with the predicate symbol \emph{@numerical} but is not yet bound to any particular value. 

In the \emph{numerical search} stage, the learner searches for values for the numerical variables using the training examples. 
We encode the search for numerical values as a satisfiability modulo theories (SMT) formula. 
For instance, to find values for $N$ in the hypothesis in Figure \ref{fig:intermediate}, the learner executes the partial hypothesis without its numerical literal \textit{geq(D,N)} against the examples to find possible substitutions for the variable $D$, from which it builds a system of linear inequations. These inequations constrain the search for the numerical variable $N$ with the values obtained for $D$ from the examples. Finally, the learner substitutes $N$ in the partial program with any solution found for the inequations. 

To implement our idea, we build on the state-of-the-art \emph{learning from failures} (LFF) \cite{cropper2020b} ILP approach.
LFF is a constraint-driven ILP approach where the goal is to accumulate constraints on the hypothesis space. A LFF learner continually generates and tests hypotheses, from which it infers constraints. 
We implement our numerical reasoning approach in \name, which, as it builds on the LFF learner \popper, supports predicate invention and learning recursive and optimal (in textual complexity) programs. \name{} uses built-in numerical literals to additionally support linear arithmetical reasoning over real numbers and integers.

\subsubsection{Novelty and Contributions}
Compared to existing approaches, the main novelty of our approach is expressivity: \name{} can learn programs with numerical values which require reasoning over multiple examples in linear arithmetic fragments.
In other words, our approach can learn programs that existing approaches cannot.
For instance, our experiments show that our approach can learn programs of the form shown in Figure \ref{fig:target}. 
In addition, our approach can (i) efficiently search for numerical values in infinite domains such as real numbers or integers, (ii) identify numerical values which may not appear in the background knowledge, and (iii) learn programs with several chained numerical literals. For instance, it can learn that the sum of two variables is lower than some particular numerical value.
As far as we are aware, no existing approach can efficiently solve such problems.
Overall, we make the following contributions:
\begin{enumerate}
\item We introduce an approach for numerical reasoning in infinite domains. 
Our approach supports numerical reasoning in linear arithmetic fragments. 
\item We implement our approach in \name, which can learn programs with numerical values, perform predicate invention, and learn recursive and optimal programs.
\item We experimentally show on four domains (geometry, biology, game playing, and program synthesis) that our approach can (i) learn programs requiring numerical reasoning, and (ii) outperform existing ILP systems in terms of learning time and predictive accuracy.
\end{enumerate}

\section{Related Work}


\paragraph{Program Synthesis.}
Program synthesis approaches that enumerate the search space \cite{raghothaman2019,ellis2018,evans2021} can only learn from small and finite domains and cannot learn from infinite domains.
Several program synthesis systems delegate the search for programs to an SMT solver \cite{jha2010,gulwani2011b,reynolds2015,albarghouthi2017}. 
By contrast, we delegate the numerical search to an SMT solver. 
Moreover, \name{} can learn programs with numerical values from infinite domains.  
Sketch \cite{solar2009} uses a SAT solver to search for suitable constants given a partial program, where the constants can be numerical values. This approach is similar to our numerical search stage. 
However, Sketch does not learn the structure of programs but expects as input a skeleton of a solution: it requires a partial program and its task is to fill in missing values with constants symbols. By contrast, \name{} learns both the program and numerical values.




\paragraph{ILP.}
Many ILP approaches \cite{muggleton1995,srinivasan2001} use bottom clause construction to search for programs.
However, these approaches can only identify numerical values that appear in the bottom clause of a single example. 
They cannot reason about multiple examples jointly, which is, for instance, necessary to learn inequations.

Constraint inductive logic programming \cite{sebag1996} uses constraint logic programming to learn programs with numerical values. 
This approach generalises a single positive example given some negative examples and is restricted to numerical reasoning in difference logic. 

\citet{anthony1997} propose an algorithm to learn hypotheses with numerical literals.
FORS \cite{karalivc1997} fits regression lines to subsets of the positive examples in a positive example only setting. In contrast to \name, these two approaches allow some error in numerical values predicted by numerical literals. 
However, these two approaches follow top-down refinement with one specialisation step at a time, which prevents them from learning hypotheses with multiple chained numerical literals.


TILDE \cite{blockeel1998} uses a discretization procedure to find candidate numerical constants while making the induction process more efficient \cite{blockeel1997}. However, TILDE cannot learn recursive programs and struggles to learn from small numbers of examples.

Many recent ILP systems enumerate every possible rule in the search space \cite{corapi2011,kaminski2018,evans2018,schuller2018} or all constant symbols as unary predicate symbols \cite{evans2018,cropper2020b,hopper} and therefore cannot handle infinite domains. 

\paragraph{LFF.} 
Recent LFF systems represent constant symbols with unary predicate symbols \cite{cropper2020b,hopper}, which prevents them from learning in infinite domains. 
\magicpopper{} \cite{hocquette2022} can identify constant symbols from infinite domains. 
Similar to \ale's lazy evaluation approach and our program search approach, \magicpopper{} builds partial hypotheses with variables in place of constant symbols.
It then executes the partial hypotheses independently over each example to identify particular candidate constant symbols. 
However, it may find an intractable number of candidate constants when testing hypotheses with non-deterministic predicates with a large or infinite number of answer substitutions, such as \emph{greater than}. 
Moreover, it cannot perform numerical reasoning from multiple examples jointly.
By contrast, \name{} uses all of the examples simultaneously when reasoning about numerical values so it can learn intervals whereas \magicpopper{} cannot.

\paragraph{Lazy Evaluation.} The most related work is an extension of \ale{} that supports \emph{lazy evaluation} \cite{srinivasan1999}. During the construction of the bottom clause, \ale{} replaces numerical values with existentially quantified variables. During the refinement search of the bottom clause, \ale{} finds substitutions for these variables by executing the partial hypothesis on the examples. This procedure can predict output numerical variables using custom loss functions measuring error \cite{srinivasan2006}, while \name{} cannot. 
However, \ale{} needs the user to write background definitions to find numerical values, such as a definition for computing a threshold or linear regression coefficients from data. 
By contrast, \name{} has numerical literals built-in.
Moreover, \ale{} executes each definition used during lazy evaluation independently which prevents it from learning hypotheses with multiple literals requiring lazy evaluation sharing variables, such as an upper and a lower bound for the same variable. By contrast, \name{} can learn hypotheses with multiple chained numerical literals. 
Finally, \ale{} does not support predicate invention, is not guaranteed to learn optimal (textually minimal) programs, and struggles to learn recursive programs.


\section{Problem Setting}
We now describe our problem setting. We assume familiarity with logic programming \cite{lloyd:book}. 
Our problem setting is the learning from failures (LFF) \cite{cropper2020b} setting, which is based on the learning from entailment setting \cite{muggleton1994} of ILP. LFF assumes a  meta-language $\cal{L}$, which is a language about hypotheses. LFF uses hypothesis constraints to restrict the hypothesis space. Hypothesis constraints are expressed in $\cal{L}$. 
A LFF input is defined as:
\begin{definition}
A LFF input is a tuple $(E^{+},E^{-},B,{\cal{H}},C)$ where $E^{+}$ and $E^{-}$ are sets of ground atoms representing positive and negative examples respectively, $B$ is a definite program representing background knowledge, ${\cal{H}}$ is a hypothesis space, and $C$ is a set of hypothesis constraints expressed in the meta-language $\cal{L}$.
\end{definition}
\noindent
Given a set of hypotheses constraints $C$, we say that a hypothesis $H$ is consistent with $C$ if, when written in ${\cal{L}}$, $H$ does not violate any constraint in $C$. We call ${\cal{H}}_{C}$ the subset of $\cal{H}$ consistent with $C$. 
We define a LFF solution:
\begin{definition}

\noindent
Given a LFF input $(E^{+},E^{-},B,{\cal{H}},C)$, a LFF solution is a hypothesis $H \in {\cal{H}}_C$ such that $H$ is complete with respect to $E^+$ ($\forall e \in E^+, B\cup H \models e$) and consistent with respect to $E^-$ ($\forall e \in E^-, B\cup H \not\models e$). \label{solution}
\end{definition}
\noindent
Conversely, given a LFF input, a hypothesis $H$ is \emph{incomplete} when $\exists e \in E^{+}, H \cup B \not\models e$, and is \emph{inconsistent} when $\exists e \in E^{-}, H \cup B \models e$.

In general, there might be multiple solutions given a LFF input. We associate a cost to each hypothesis and prefer \textit{optimal} solutions, which are solutions with minimal cost. In the following, we use as cost function the size of hypotheses, measured as the number of literals in it. 

A hypothesis which is not a solution is called a failure. A LFF learner identifies constraints from failures to restrict the hypothesis space. For instance, if a hypothesis is inconsistent, a \emph{generalisation} constraint prunes its generalisations, as they are provably also inconsistent.
\section{Numerical Reasoning}
%

We extend the framework in the previous section to allow numerical reasoning in possibly infinite domains. We assume familiarity with SMT theory \cite{smt}. The idea is to separate the search into two stages (i) \emph{program search}, and (ii) \emph{numerical search}. First, the learner generates partial programs with first-order numerical variables in place of numerical values. Then, the learner searches for numerical values to fill in the numerical variables.
\subsection{Program Search}
The learner first searches for partial programs with variables, called numerical variables, in place of numerical values.
\paragraph{Numerical Variables.} We extend the meta-language $\cal{L}$ of LFF to contain numerical variables. A numerical variable is a first-order variable that can be substituted by a numerical value, i.e. a numerical variable acts as a placeholder for a numerical symbol.
In the following, we represent numerical variables with the unary predicate symbol \emph{@numerical}. 
For example, in the program in Figure \ref{fig:intermediate}, the variable $N$ marked with the syntax \emph{@numerical} is a numerical variable.
\paragraph{Numerical Literals.} 
A numerical literal is a literal which requires numerical reasoning and whose arguments all are numerical. A numerical literal may contain numerical variables as arguments.
During the \emph{program search} stage, the learner builds partial hypotheses with variables in place of numerical variables in numerical literals. For example, the learner may generate the following program, where the literal \textit{leq(B,N)} is a numerical literal which contains the numerical variable $N$:
\begin{center}
\begin{tabular}{l}
\emph{H: f(A) $\leftarrow$ length(A,B), leq(B,N), @numerical(N)}\\
\end{tabular}
\end{center}

\paragraph{Related Variables.} 
A related variable is a variable that appears both in a numerical literal and a regular literal. 
Related variables act as bridges between relational learning and numerical reasoning. 
For instance, the variable $B$ is a variable related to the numerical variable $N$ in the program $H$ above. Possible substitutions for the related variables are identified by executing the hypothesis without its numerical literals over the positive and negative examples.
For instance, given the positive examples $\{f([a,b]), f([])\}$ 
and the negative examples $\{f([b,c,a,d,e,f]), f([c,e,d,a,b])\}$, the hypothesis $H$ above has the following positive $S_P(B)$ and negative $S_N(B)$ substitutions for the related variable $B$: $S_P(B) = \{2,0\}$ and $S_N(B)=\{6,5\}$.
\subsection{Numerical Search} \label{numerical_search}
During the numerical search stage, the learner builds an SMT formula from the definition of the numerical literals and the possible substitutions for the related variables. 
It generates a constraint for each positive example to ensure the learned hypothesis covers it. It generates a constraint for each negative example to ensure the learned hypothesis does not cover it.
For instance, the learner translates the numerical search in the hypothesis above as the following SMT formula:
\begin{align*}
2 \leq N \land 0 \leq N \land \neg(6 \leq N) \land \neg(5 \leq N)
\end{align*}
 The appendix includes details of how \name{} builds the SMT formula.
 The solutions for the formula represent possible numerical values for the partial program tested. In other words, if the formula is satisfiable, any solution is a substitution for the numerical variables in the partial program such that the resulting program is a solution to the LFF input. For instance, the substitution $N=3$ is a solution to the formula above. Applying this substitution to the program $H$ above forms the following LFF solution:
\begin{center}
\begin{tabular}{l}
\emph{H: f(A) $\leftarrow$ length(A,B), leq(B,3)}\\
\end{tabular}
\end{center}
In practice, to account for non-deterministic literals, we build one expression from each of the substitutions found for the related variables. 
The constraints assert that at least one of these expressions is verified for each positive example and none are verified for any negative examples. 
In other words, the multi-instance problem \cite{dietterich1997} is delegated to the solver through a disjunction.

The number of literals in the resulting SMT formula is upper bounded by $n_e*s*n_v$, where $n_e$ is the number of examples,  $s$ is the maximum number of substitutions per example, and $n_v$ is the number of variables in the candidate hypothesis. A proof of this result is in the appendix.
\subsection{Constraints}
If a candidate program is not a solution to the LFF input, we generate constraints to prune other programs from the hypothesis space and constrain subsequent \emph{program search} stages. Following \citet{hocquette2022}, we use the following constraints. Given a partial program $P$ with numerical variables generated in the \emph{program search} stage:
\begin{enumerate}
    \item If there is no solution in the \emph{numerical search} stage, then $P$ cannot cover any of the positive examples and therefore $P$ is too specific. We prune programs which include one of the specialisations of $P$ as a subset.
    \item If all solutions found in the \emph{numerical search} stage result in programs which are too specific, then $P$ is too specific. We prune specialisations of $P$ without additional numerical literals.
    \item If all solutions found in the \emph{numerical search} stage result in programs which are too general, then $P$ is too general. We prune non-recursive generalisations of $P$.
\end{enumerate}
These constraints are optimally sound \cite{hocquette2022} as they do not prune optimal solutions from the hypothesis space. 
The appendix contains more details about the constraints.
\section{Implementation}
We present our implementation called \name. We first briefly describe \popper{} \cite{cropper2020b}, on which \name{} is based. 
\subsection{\popper}
\popper{} takes as input a LFF input, which contains a set of positive and negative examples, a background knowledge $B$, a bound over the size of hypotheses allowed in ${\cal{H}}$, and a set of hypothesis constraints $C$. To generate hypotheses, \popper{} uses an ASP program $P$ whose models are hypothesis solutions represented in the meta-language $\cal{L}$. In other words, each model (answer set) of $P$ represents a hypothesis. \popper{} follows a \textit{generate}, \textit{test}, and \textit{constrain} loop to find a solution. First, it generates a hypothesis as a solution to the ASP program $P$ with the ASP system Clingo \cite{gebser2014}. \popper{} tests this hypothesis given the background knowledge against the examples, typically using Prolog. If the hypothesis is a solution, \popper{} returns it. Otherwise, the hypothesis is a failure: \popper{} identifies the kind of failure and builds constraints accordingly. For instance, if the hypothesis is inconsistent, \popper{} builds a generalisation constraint. 
\popper{} adds these constraints to the ASP program $P$ to constrain the subsequent \textit{generate} steps. 
This loops repeats until a hypothesis solution is found or until there are no more models to the ASP program $P$.

\subsection{\name} \name{} builds on \popper. It also follows a \textit{generate}, \textit{test}, and \textit{constrain} loop.

\begin{figure}[t]
\footnotesize
\centering
\subfloat{
\begin{tabular}{@{}lll@{}}
\footnotesize
\textbf{Literal} & \textbf{Definition} & \textbf{Example}\\\hline 
\texttt{geq(A,N)} & \texttt{$A \geq N$} & \texttt{geq(A,3)}\\
\texttt{leq(A,N)} & \texttt{$A \leq N$} & \texttt{leq(A,5.2)}\\
\texttt{add(A,B,C)} & \texttt{$A+B=C$} & \texttt{add(A,B,C)}\\
\texttt{mult(A,N,C)} & \texttt{$A*N=C$} & \texttt{mult(A,2,C)}\\ 
\end{tabular}
}
    \caption{Numerical literals in \name{}. $N$ is a numerical variable which can be substituted for a numerical value. Variables $A$, $B$, $C$, $N$ range over real numbers or integers.}
    \label{fig:builtinpreds}
\end{figure}

\begin{figure}[t]
\footnotesize
\subfloat{
\resizebox{0.48\textwidth}{!}{ 
\begin{tabular}{@{}lcl@{}}
\textbf{Fragment} & \textbf{\name} & \textbf{Example}\\\hline 
Linear Real Arithmetic & $\checkmark$ & $X+6.3*Y \leq 3$\\
Linear Integer Arithmetic & $\checkmark$ & $A+6*B \leq 3$\\
Mixed Real / Integer & $\checkmark$ & $X+6.3*B \leq 3$\\
Integer Difference Logic & $\checkmark$ & $A-B \leq 4$\\
Real Difference Logic & $\checkmark$ & $X-Y \leq 4$\\
Unit two-variable / inequality & $\checkmark$ & $X+Y \leq 4$\\
Polynomial Real Arithmetic & \text{\sffamily X} & $X^2+Y^2=2$\\
Non-linear Integer Arithmetic & \text{\sffamily X} & $A^2 = 2$\\ 
\end{tabular}
}
}
    \caption{Arithmetical fragments supported by \name. $X$ and $Y$ range over real numbers and $A$ and $B$ over integers.
    }
    \label{fig:fragments}
\end{figure}
\paragraph{Partial programs.} First, \name{} generates partial programs which may contain numerical literals. 
The maximum number of numerical literals in a clause is a user parameter, with a default value of 2. 
This setting expresses the trade-off between search complexity and expressivity.
\paragraph{Numerical literals.} 
\name{} supports the built-in numerical literals shown in Figure \ref{fig:builtinpreds}. 
While \textit{add} reasons from regular numerical first-order variables and does not have numerical variables which are substituted for constant symbols, other literals have such numerical variables represented by $N$. As shown in Figure \ref{fig:fragments}, the numerical literals in \name{} are sufficient to reason about standard linear arithmetic fragments. Other fragments which currently are not supported by \name{} include non-linear arithmetic for complexity reasons. More details about numerical literals are provided in the appendix.
The user can specify a subset of these numerical literals to use if this bias is known. Otherwise, \name{} automatically identifies which of these literals to use, at the expense of more search. If known, the user also can optionally specify argument types (real or integer) and domains for the numerical variables to restrict the search. 
\paragraph{Numerical reasoning.} 
\name{} performs numerical reasoning during the \textit{test} stage. 
It first identifies possible substitutions for the related variables. To do so, it adds related variables in numerical literals as new arguments to the head literal. Then, it removes numerical literals from the hypothesis. For instance, the hypothesis $H$ below becomes $H'$:
\begin{center}
\begin{tabular}{l}
\emph{H: f(A) $\leftarrow$ length(A,B), leq(B,C)}\\
\emph{H': f(A,B) $\leftarrow$ length(A,B)}\\
\end{tabular}
\end{center}
\name{} executes the resulting hypothesis over the examples with Prolog. 
We use Prolog because of its ability to handle lists and large, potentially infinite, domains. \name{} saves the  substitutions found for the newly added head variables. It then builds an SMT formula from the definition of the numerical literals and the values found for the related variables. Finally, \name{} uses the SMT solver Z3 \cite{moura2008} to determine the satisfiability of the resulting SMT formula. 
If a solution exists, it saves a possible value for each numerical variable and substitutes these values into the original program. Otherwise, it repeats the loop to generate more programs. 

We set the SMT solver to return any solution to the formula.
We do not optimise the choice of numerical values because it is unclear how to trade off learning textually minimal programs and learning optimal numerical values (potentially multiple ones in a program).
Addressing this limitation is future work.


\section{Experiments}
We claim that \name{} can learn programs with numerical values from numerical reasoning. Therefore, our experiments aim to answer the following question:
\begin{enumerate}
\item[\textbf{Q1}] Can \name{} learn programs with numerical values?
\end{enumerate}
To answer \textbf{Q1}, we evaluate \name{} on a variety of tasks requiring numerical reasoning. 

We also claim that our approach can reduce search complexity and thus improve learning performance. Therefore, our experiments aim to answer the following question:
\begin{enumerate}
\item[\textbf{Q2}] How well does \name{} perform compared to other approaches?
\end{enumerate}
To answer \textbf{Q2}, we compare \name{} against the systems \ale{} and \magicpopper{}, which are the only program synthesis systems capable of learning programs with numerical constant symbols\footnote{We also considered other systems \cite{corapi2011,evans2018,kaminski2018,schuller2018}. However, these systems cannot handle infinite domains and therefore cannot solve any of the tasks proposed or require user-provided metarules \cite{muggleton2015} making them unusable in practice.
}.

As described in Section \ref{numerical_search}, the size of the SMT formula built by \name{} is an increasing function of the number of examples. Therefore, to evaluate how well our system scales, we investigate the following question:
\begin{enumerate}
\item[\textbf{Q3}] How well does \name{} scale with the number of examples?
\end{enumerate}
To answer \textbf{Q3}, we vary the number of examples and evaluate the performance of \name.


\subsubsection{Domains}
We consider four domains to evaluate our claims. 
The appendix includes more details about the domains and tasks.

\textbf{Geometry.}
These tasks involve learning that points belong to geometrical objects (interval, half-plane), which parameters are numerical values to be learned. 
Figure \ref{fig:half_plane} shows an example hypothesis for the task \emph{half-plane}.
 \begin{figure}
 \subfloat{
 \footnotesize
\begin{tabular}{l}
\texttt{halfplane(A,B) $\leftarrow$ \textbf{mult}(A,\textbf{3},D), \textbf{add}(B,D,E), \textbf{leq}(E,\textbf{6}).}\\
\end{tabular}
}
    \caption{Example \emph{halfplane} hypothesis. Numerical literals and examples of numerical values are in bold.
    }\label{fig:half_plane}
\end{figure}

 \begin{figure}
 \subfloat{
 \footnotesize
\begin{tabular}{l}
\texttt{zendo2(A) $\leftarrow$ piece(A,B), position(B,C,D),}\\ \hspace{56pt} \texttt{\textbf{add}(C,D,E), \textbf{leq}(E,\textbf{6.92}).}\\
\texttt{zendo2(A) $\leftarrow$ piece(A,B), rotation(B,D),}\\
\hspace{56pt} \texttt{\textbf{leq}(D,\textbf{4.12}), \textbf{geq}(D,\textbf{3.23})}\\
\end{tabular}
}
    \caption{Example \emph{zendo2} hypothesis. Numerical literals and examples of numerical values are in bold.
    }\label{fig:zendo2}
\end{figure}
\textbf{Zendo.}
Zendo is a multiplayer game in which players aim to identify a rule which structures made from a set of pieces with varying attributes must follow. 
We consider four increasingly complex tasks. Figures \ref{fig:target} and \ref{fig:zendo2} show examples of target hypotheses for tasks 1 and 2, respectively. 

\textbf{Pharmacophores.}
The goal is to identify properties of pharmacophores responsible for medicinal activity \cite{finn1998}.
This domain requires reasoning about distances between atoms with varying properties and bonds linking each other. 
 We consider four increasingly complex tasks. Figure \ref{fig:pharma} shows an example of a target hypothesis for task 4. 

 \begin{figure}
 \subfloat{
 \footnotesize
\begin{tabular}{l}
\texttt{pharma4(A) $\leftarrow$ zinc(A,B), hacc(A,C), dist(A,B,C,D),} \\ \hspace{60pt}
 \texttt{\textbf{leq}(D,\textbf{4.18}), \textbf{geq}(D,\textbf{2.22})}.\\
\texttt{pharma4(A) $\leftarrow$ hacc(A,C), hacc(A,E), dist(A,B,C,D),}\\ \hspace{60pt} \texttt{\textbf{geq}(D,\textbf{1.23}), \textbf{leq}(D,\textbf{3.41})}.\\
\texttt{pharma4(A)  $\leftarrow$ zinc(A,C), zinc(A,B), bond(B,C,du),}\\ \hspace{60pt}
\texttt{dist(A,B,C,D), \textbf{leq}(D,\textbf{1.23})}.
\end{tabular}
}
    \caption{Example \emph{pharma4} hypothesis. Numerical literals and examples of numerical values are in bold.
    }\label{fig:pharma}
\end{figure}

\textbf{Program Synthesis.}
We consider three program synthesis tasks. These tasks are list transformation tasks which involve learning recursive programs and numerical reasoning. 
\subsubsection{Systems}
To evaluate \textbf{Q2}, we compare \name{} against \magicpopper{} and \ale{}. We briefly describe each of these systems. 
The appendix contains more details.

\textbf{\magicpopper{} and \name.} 
 We provide \name{} and \magicpopper{} with the same input. We allow \magicpopper{} to learn constant symbols for variables of type real or integer in literals with a finite number of answer substitutions. 
The experimental difference is the ability to perform numerical reasoning for \name. 

\textbf{\ale.} 
We provide \ale{} with definitions adapted from \name's numerical literals to fit its lazy evaluation procedure.
\ale{} uses a different bias than \name{} to bound the hypothesis space. Therefore, the comparison is less fair and should be interpreted as indicative only. 
\subsubsection{Experimental Setup}
We enforce a timeout of 10 minutes per task. We measure predictive accuracy and learning time. We measure the mean and standard error over 10 trials. We use an 8-Core 3.2 GHz Apple M1 and a single CPU\footnote{
The code and experimental data for reproducing the experiments are provided as supplementary material and will be made publicly available if the paper is accepted for publication.
}.
\subsection{Experiment 1: Comparison Against SOTA}
\begin{table}[ht]
\centering
\footnotesize
\begin{tabular}{@{}l|ccc@{}}
\textbf{Task} & \textbf{\ale} & \textbf{\magicpopper} & \textbf{\name}\\
\midrule
\emph{interval} & 69 $\pm$ 1 & 70 $\pm$ 0 & \textbf{99 $\pm$ 1}\\
\emph{halfplane} & \textbf{99 $\pm$ 0} & 84 $\pm$ 7 & 96 $\pm$ 1\\\midrule
\emph{zendo1} & 98 $\pm$ 0 & 68 $\pm$ 3 & \textbf{99 $\pm$ 0}\\ 
\emph{zendo2} & 51 $\pm$ 1 & 56 $\pm$ 1 & \textbf{96 $\pm$ 1}\\ 
\emph{zendo3} & 71 $\pm$ 1 & 51 $\pm$ 1 & \textbf{96 $\pm$ 1}\\
\emph{zendo4} & 63 $\pm$ 1 & 52 $\pm$ 1 & \textbf{94 $\pm$ 1}\\\midrule
\emph{pharma1} & 82 $\pm$ 1 & 64 $\pm$ 3 & \textbf{99 $\pm$ 0}\\
\emph{pharma2} & 83 $\pm$ 1 & 77 $\pm$ 2 & \textbf{95 $\pm$ 1}\\
\emph{pharma3} & 81 $\pm$ 1 & 82 $\pm$ 1 & \textbf{98 $\pm$ 1}\\
\emph{pharma4} & 76 $\pm$ 1 & 62 $\pm$ 2 & \textbf{92 $\pm$ 1}\\\midrule
\emph{member$\_$between} & 49 $\pm$ 0 & 75 $\pm$ 4 & \textbf{97 $\pm$ 1}\\
\emph{last$\_$leq} & 50 $\pm$ 0 & 51 $\pm$ 1 & \textbf{98 $\pm$ 1}\\
\emph{next$\_$geq} & 50 $\pm$ 0 & 50 $\pm$ 0 & \textbf{92 $\pm$ 5}\\
\end{tabular}
\caption{
Predictive accuracies. We round accuracies to integer values. The error is standard error.
}
\label{tab:accuracies}
\end{table}

\begin{table}[ht]
\centering
\footnotesize
\begin{tabular}{@{}l|ccc@{}}
\textbf{Task} & \textbf{\ale} & \textbf{\magicpopper} & \textbf{\name}\\
\midrule
\emph{interval} & 1 $\pm$ 0 & \textbf{0 $\pm$ 0} & \textbf{0 $\pm$ 0}\\
\emph{halfplane} & \textbf{1 $\pm$ 0} & 60 $\pm$ 26 & 2 $\pm$ 1\\\midrule
\emph{zendo1} & 25 $\pm$ 8 & timeout & \textbf{10 $\pm$ 1}\\
\emph{zendo2} & 68 $\pm$ 18 & 97 $\pm$ 11 & \textbf{17 $\pm$ 1}\\
\emph{zendo3} & 106 $\pm$ 26 & 112 $\pm$ 9 & \textbf{69 $\pm$ 2}\\
\emph{zendo4} & 147 $\pm$ 30 & timeout & \textbf{76 $\pm$ 2} \\\midrule
\emph{pharma1} & 2 $\pm$ 0 & 3 $\pm$ 0 & \textbf{1 $\pm$ 0}\\
\emph{pharma2} & 10 $\pm$ 2 & 7 $\pm$ 0 & \textbf{2 $\pm$ 0}\\
\emph{pharma3} & 24 $\pm$ 3 & 66 $\pm$ 5 & \textbf{20 $\pm$ 1}\\
\emph{pharma4} & \textbf{3 $\pm$ 0} & 62 $\pm$ 2 & 20 $\pm$ 0\\\midrule
\emph{member$\_$between} & \textbf{1 $\pm$ 0} & 161 $\pm$ 38 & 2 $\pm$ 0\\
\emph{last$\_$leq} & \textbf{0 $\pm$ 0} & 589 $\pm$ 10 & 13 $\pm$ 1\\
\emph{next$\_$geq} & \textbf{0 $\pm$ 0} & 336 $\pm$ 17 & 39 $\pm$ 6\\
\end{tabular}
\caption{
Learning times. We round times over one second to the nearest second. The error is standard error.}
\label{tab:time}
\end{table}

Tables \ref{tab:accuracies} and \ref{tab:time} show the results. They show that \name{} consistently achieves high accuracy on all tasks. The accuracy is not maximal because, given a training set, several numerical values may result in a complete and consistent hypothesis, and \name{} does not optimise the choice of numerical values. For instance, given the SMT formula, $2 \leq N \land 0 \leq N \land \neg(6 \leq N) \land \neg(5 \leq N)$, \name{} could return any value $N$ such that $2 \leq N < 5$. 

These results demonstrate that \name{} can learn programs with numerical values in a reasonable time (less than 80s) in a variety of domains. \name{} can identify numerical values which require reasoning from multiple examples and which may not appear in the background knowledge. For instance, it can solve \emph{pharma1} which involves learning that the distance between two atoms must be smaller than a particular value.
It can also learn programs with numerical values from infinite domains, such as real numbers or integers. Finally, it can learn hypotheses with multiple related numerical literals for instance for \emph{halfplane} or \emph{pharma4} (Figures \ref{fig:half_plane} and \ref{fig:pharma}). Given these results, we answer \textbf{Q1} positively.

We compare \name{} against \ale{} and \magicpopper{}. Table \ref{tab:accuracies} shows \name{} achieves higher or equal accuracies than both \ale{} and \magicpopper. An independent t-test confirms the significance of the difference at the $p < 0.01$ level for all tasks except \emph{halfplane} and \emph{zendo1}. These results show that \name{} can solve tasks other systems struggle with. For instance, \ale{} struggles to learn hypotheses with multiple numerical literals sharing variables which is necessary for instance for \textit{zendo2} or \textit{zendo3}. 
\ale{} performs lazy evaluation over all substitutions for positive and negative examples and, therefore, struggles to learn numerical hypotheses with disjunctions such as for \textit{zendo2} or \textit{pharma2}. 
In these situations, \ale{} may learn a hypothesis as facts, which do not generalise to the test set.
Finally, \ale{} struggles to learn recursive programs and performs poorly on the program synthesis tasks. Otherwise, \ale{} performs well on other tasks, such as \textit{halfplane} or \textit{zendo1}.

\magicpopper{} can learn programs, potentially recursive ones, with constant symbols from infinite domains. 
However, it cannot reason from multiple examples jointly and cannot identify constants in literals with large or infinite number of substitutions, such as \emph{greater than}. These limitations prevent it from learning inequalities, such as in \emph{pharma2}. 

Table \ref{tab:time} shows the learning times. It shows \ale{} can have longer learning times than \name{}. For instance, \ale{} solves \textit{zendo1} in 25s while \name{} requires 10s.
Yet, in contrast to \ale, \name{} searches for textually optimal solutions. 
\name{} also outperforms \magicpopper{} in terms of learning times. Owing to the lack of numerical reasoning ability, \magicpopper{} is unable to express a concise hypothesis on some tasks and therefore searches up to larger depth. Moreover, \magicpopper{} follows a generate-and-test approach to identify numerical values: it first generates candidate numerical values from the execution of the partial programs over single examples. Then, it tests candidate values over all examples. Conversely, \name{} solves a single problem with all examples jointly. It thus avoids the need to consider possibly many candidate values and can be more efficient. Given these results, the answer to \textbf{Q2} is that \name{} can outperform existing approaches in terms of learning times and predictive accuracies when learning programs with numerical values.


\subsection{Experiment 2: Scalability}
We compare the performance of \name{} against \ale{} and \magicpopper{} when varying the number of training examples. We use the task \textit{zendo1}, which other systems can solve. However, the main advantage of our approach is that it can learn concepts existing approaches cannot. 
Therefore, we also evaluate scalability on the task \textit{pharma2}, which existing systems struggle to solve.
Figures \ref{fig:timezendo1} and \ref{fig:pharm_time} show the learning times versus the number of examples. The appendix shows the predictive accuracies. They are not maximal for \magicpopper{} and \ale{} on \emph{pharma2}.


\definecolor{mygreen}{rgb}{0.1,0.6,0.1}
\pgfplotstableread[col sep=comma]{./zendoexamples/26_7_zendo_1__aleph.csv}\resaleph
\pgfplotstableread[col sep=comma]{./figures/new_combine/Zendo/magicpopper.csv}\resmagicpopper
\pgfplotstableread[col sep=comma]{./figures/new_combine/Zendo/numericalpopper.csv}\resnumericalpopper

\pgfplotstableread[col sep=comma]{./figures/new_combine/pharma/numericalpopper.csv}\resnumericalpopperpharma
\pgfplotstableread[col sep=comma]{./figures/new_combine/pharma/magicpopper.csv}\resmagicpopperpharma
\pgfplotstableread[col sep=comma]{./pharmaexamples/pharma2__aleph.csv}\resalephpharma


\begin{figure}
  \begin{minipage}{0.05\textwidth}
\begin{tikzpicture}
\begin{customlegend}[legend columns=5,legend style={nodes={scale=1, transform shape},align=left,column sep=0ex},
        legend entries={\name, \ale, \magicpopper}]
        \addlegendimage{red,mark=diamond*}
        \addlegendimage{blue,mark=square*}
        \addlegendimage{mygreen,mark=triangle*}
\end{customlegend}
\end{tikzpicture}
\end{minipage}\hfill\\
  \begin{minipage}{0.22\textwidth}
\resizebox{\columnwidth}{!}{
\begin{tikzpicture}
\begin{axis}[
  legend style={at={(0.5,0.35)},anchor=west},
   legend style={font=\normalsize},
  tick label style={font=\Large},
  xlabel style={font=\Large},
  ylabel style={font=\Large},
  xtick={0,100,200,300,400,500},
  ytick={0,100,200, 300, 400, 500, 600},
  xlabel=Number of examples,
  ylabel=Learning time (s),
  xmin=0,
  xmax=500,
  ymin=0,
  ymax=610,
  ]
\addplot+[blue,mark=square*,
                error bars/.cd,
                y dir=both,
                error mark,
                y explicit]table[x=xs,y=time_av,y error=time_std] {\resaleph};
\addplot[red,mark=diamond*,
                error bars/.cd,
                y dir=both,
                error mark,
                y explicit]table[x=xs,y=time_av,
    y error=time_std] {\resnumericalpopper};
\addplot[mygreen,mark=triangle*,
                error bars/.cd,
                y dir=both,
                error mark,
                y explicit]table[x=xs, y=time_av,
    y error=time_std] {\resmagicpopper};
\end{axis}
\end{tikzpicture}}
\caption{Learning time versus the number of examples for \emph{zendo1}.}
\label{fig:timezendo1}
\end{minipage}\hfill
  \begin{minipage}{0.22\textwidth}
\resizebox{\columnwidth}{!}{
\begin{tikzpicture}
\begin{axis}[
  legend style={at={(0.5,0.35)},anchor=west},
   legend style={font=\normalsize},
  tick label style={font=\Large},
  xlabel style={font=\Large},
  ylabel style={font=\Large},
  xtick={0,100,200,300,400,500},
  ytick={0,100,200,300,400,500,600},
  xlabel=Number of examples,
  ylabel=Learning time (s),
  xmin=0,
  xmax=500,
  ymin=0,
  ymax=610,
  ]
\addplot[red,mark=diamond*,
                error bars/.cd,
                y dir=both,
                error mark,
                y explicit]table[x=xs,y=time_av,
    y error=time_std] {\resnumericalpopperpharma};
    \addplot[mygreen,mark=triangle*,
                error bars/.cd,
                y dir=both,
                error mark,
                y explicit]table[x=xs,y=time_av,
    y error=time_std] {\resmagicpopperpharma};
    \addplot[blue,mark=square*,
                error bars/.cd,
                y dir=both,
                error mark,
                y explicit]table[x=xs,y=time_av,
    y error=time_std] {\resalephpharma};
\end{axis}
\end{tikzpicture}}
\caption{Learning time versus the number of examples for \emph{pharma2}.}
\label{fig:pharm_time}
\end{minipage}
\end{figure}

As the number of examples grows, 
the complexity of the learning task, and thus the learning time, increases. Predictive accuracies degrade when learning times reach timeout (i.e. for \ale{} and \magicpopper). \name{} has shorter learning times than \magicpopper{} on both tasks and its learning time increases slower. \magicpopper{} generates all candidate numerical values derivable from single examples, then tests them against the remaining examples. Conversely, \name{} generates constraints from all examples before solving the SMT formula. It thus can propagate information earlier on which results in shorter learning times. 
However, owing to the complexity of the SMT problem, \name{} can struggle to scale to large numbers of examples. For instance, the SMT formula can include disjunctions in the case of non-deterministic literals which further adds complexity. The appendix includes a breakdown of the learning times of \name{}. It shows its learning time is dominated by the construction and solving of the SMT formula. Finally, \name{} scales better than \ale{} on \emph{pharma2} but worse on \emph{zendo1}. Therefore, the answer to \textbf{Q3} is that \name{} scales better than \magicpopper{} with the number of examples and can scale better than \ale. However, scalability is limited by the complexity of the numerical reasoning stage. This result highlights one limitation of \name.
\section{Conclusions and Future Work}
Learning programs with numerical values is essential for many AI applications. 
However, existing program synthesis systems struggle to identify numerical values from infinite domains and reason about multiple examples. 
To overcome these limitations, we have introduced \name, an ILP system that combines relational learning and numerical reasoning to efficiently learn programs with numerical values. 
The key idea of our approach is to decompose learning into two stages: (i) the search for a program, and (ii) the search for numerical values. 
During the search for a program, \name{} builds partial programs with variables in place of numerical values. 
Then, given a partial program, \name{} searches for numerical values by building an SMT formula using the training examples.
\name{} uses a set of built-in numerical literals (Figure \ref{fig:builtinpreds}) to support a large class of arithmetical fragments (Figure \ref{fig:fragments}), such as \emph{linear integer arithmetic}.
Our experiments on four domains (geometry, game playing, biology, and program synthesis) show that our approach can (i) learn programs with numerical values, and (ii) improve predictive accuracies and reduce learning times compared to state-of-the-art ILP systems. 
In particular, it can learn programs with multiple numerical values, including recursive programs.
In other words, we have shown that \name{} can solve numerical tasks that existing systems cannot.
At a higher level, we think that this paper helps bridge relational and numerical learning.



\subsection*{Limitations and Future Work}


\textbf{Scalability.} A limitation to the scalability of our approach is the complexity of the numerical reasoning stage, which is a function of the number of (i) examples, and (ii) numerical variables. 
Future work will aim to identify a subset of the examples which are sufficient to identify suitable numerical values \cite{anthony1997}.

\textbf{Cost Function.} \name{} learns optimal programs, where the cost function is the size of the hypothesis (the number of literals in it).
However, it might be desirable to prefer hypotheses based on different criteria, such as maximum margin or mean square error in the case of numerical prediction \cite{srinivasan1999}. Future work should explore learning with alternative cost functions.

\textbf{Noise.} In contrast to other ILP systems \cite{karalivc1997,blockeel1998,srinivasan2001}, \name{} cannot identify numerical values from noisy examples. 
\citet{wahlig2022} extended LFF to support learning from noisy examples. 
This extension should be directly applicable to \name.


\appendix

\appendixpage

This appendix contains supplementary material for the paper \emph{Relational program synthesis with numerical reasoning}.
Its outline is as follows.

\begin{itemize}
\item Appendix \ref{numericalsearch} provides details about the encoding of the numerical search stage, and a proposition providing a bound over the number of literals in the SMT formula generated by \name.
\item Appendix \ref{constrain} describes the constraints used in \name{} and an example.
\item Appendix \ref{implementation} provides details on the implementation of \name.
\item Appendix \ref{exp} describes our experiments, including sample solutions. It also provides additional results, namely the accuracy versus the number of examples for \emph{zendo1} and for \emph{pharma2} and the average proportion of learning time spent in the different learning stages for all tasks.
\end{itemize}

\section{Numerical Search} \label{numericalsearch}.
We describe in this section the \textit{numerical search} stage of \name. We assume a partial program $H$ has been identified in the \emph{program search} stage.

\name{} is based on \popperplus{} \cite{popper+}. \popperplus{} maintains a set of candidate programs which are partially complete (i.e. cover some of the positive examples) and consistent (i.e. cover no negative examples). \popperplus{} then combines some of these candidate programs into a program which is complete (i.e. covers all of the positive examples) and consistent. Therefore, we look for candidate programs which are (i) partially complete, and (ii) consistent. To ensure programs are optimal (textually minimal), we additionally look for programs which (iii) cover the maximum number of positive examples. 

During the \textit{numerical search} stage, \name{} builds an SMT formula, such that any solution, once substituted in the partial program, results in a program which verifies the conditions (i), (ii), and (iii) above. We now describe the SMT formula construction.

Each clause in the partial program $H$ is a conjunction of numerical literals and regular literals. \name{} builds the SMT formula from the numerical literals. Each numerical literal represents a relation $R$. Built-in relations $R$ are part of $\{=, \leq, \geq, +, * \}$. We call $xRy$ the expression built from the variables or constants $x$ and $y$ and the relation $R$. For instance, if $R$ is $\leq$, $x$ is the variable $v$ and $y=3$, then $xRy$ means $v \leq 3$.

Each numerical literal may contain a numerical variable. For each numerical variable, \name{} generates a variable $v$. \name{} assigns this variable $v$ the type and domain provided by the user, with type \textit{real} by default. 

For each example $e_{i}$, we call $s_{i,m,k}$ the $k^{th}$ possible substitution for the related variables in the $m^{th}$ numerical literal. Each positive example must have at least one substitution that fulfills the hypothesis $H$. Therefore, each positive example must have at least one substitution that fulfills the conjunction of numerical literals in $H$. Then, for each positive example, \name{} generates a Boolean variable $b_i$ and an expression of the form:
\begin{align*}
b_i == \lor_{k} \land_{m} s_{i,m,k} R_{m} v_{m}
\end{align*}
To ensure the maximum number of positive examples are covered, \name{} maximises the objective:
\begin{align*}
\max \sum_i b_i
\end{align*}
\name{} generates a constraint to cover at least one positive example:
\begin{align*}
\sum_{i} b_i \geq 1
\end{align*}
Each negative example must have no substitution which fulfills the hypothesis. Therefore, each negative example must have no substitution which fulfills the conjunction of numerical literals in $H$. For each negative example $e_i$, \name{} generates a constraint of the form below to ensure it is not covered:
\begin{align*}
\mathrm{not}(\lor_{k} \land_{m} s_{i,m,k} R_{m} v_{m})
\end{align*}
If a solution for this formula exists, \name{} saves an optimal solution. Then, \name{} adds a constraint specifying that at least one positive example uncovered by the solution found must now be covered. It searches for another solution covering the maximum number of positive examples that are not already covered. It repeats this process to obtain a set of candidate solutions, each covering at least one uncovered example. 

To illustrate the SMT formula construction, we provide the following example:
\begin{example}
Assume \name{} generates the following hypothesis:
\begin{center}
\texttt{zendo(A):-piece(A,B), contact(B,C),}
\texttt{size(C,D), geq(D,E), @numerical(E).}
\end{center}
First, \name{} executes the hypothesis over the examples. Assume it finds the following substitutions for the related variable $D$ from the positive and negative examples:
\begin{align*}
S_P(D) = \{[8.2, 9.4], [2.3, 10.3]\}\\
S_N(D) = \{[2.4, 4.6], [5.3, 1.2] \}
\end{align*}
In both $S_P(D)$ and $S_N(D)$, the first list corresponds to the first example and the second to the second example. Each example has two substitutions for the variable $D$.
\name{} generates a variable $v_E$ for the numerical variable $E$. \name{} generates a Boolean variable for each positive example:
\begin{align*}
b_1 == (8.2 >= v_E) \lor (9.4 >= v_E)\\ 
b_2 == (2.3 >= v_E) \lor (10.3 >= v_E)\\ 
\end{align*}
\name{} maximises the number of positive examples covered:
\begin{align*}
\max \sum_{i} b_i
\end{align*}
\name{} generates a constraint to cover at least one positive example:
\begin{align*}
\sum_{i} b_i \geq 1
\end{align*}
For each negative example, \name{} generates a constraint to ensure it is not covered:
\begin{align*}
\mathrm{not}((2.4 >= v_E) \lor (4.6 >= v_E))\\
\mathrm{not}((5.3 >= v_E) \lor (1.2 >= v_E))\\ 
\end{align*}
Finally, \name{} solves the resulting SMT formula. For this example, the two positive examples can be covered and any solution such that $5.3 < v_B <= 8.2$ may be returned.


\end{example}

\begin{proposition} Assume $n_e$ is the number of training examples, $s$ is the maximum number of substitutions per example, and $n_v$ is the number of variables in the candidate hypothesis, the number of literals in the SMT formula generated \name{} is at most $n_e*s*n_v$.
\end{proposition}
\begin{proof}
For each example $e$, \name{} generates one formula. This formula is a disjunction of $s_e \leq s$ terms, one per possible substitution for the example $e$. Each of these terms is a conjunction of $n_v$ literals, one per variable in the candidate hypothesis. Therefore, the resulting formula has at most $n_e*s*n_v$ literals.
\end{proof}
\section{Constraints} \label{constrain}
 \name{} is based on the LFF \cite{cropper2020b} setting of ILP. A LFF learner accumulates constraints to restrict the hypothesis space. It repeatedly generates hypotheses. If a hypothesis is not a solution, it generates constraints to guide future program generation. Following \cite{hocquette2022}, we add constraints which apply to partial programs with variables in place of constant symbols. For instance, if a partial program $H$ has no substitutions for its numerical variables such that the resulting hypothesis (i) covers no negative examples (ii) covers at least one positive example, we add a specialisation constraint which prunes specialisation of $H$ without additional numerical literals. 
 \begin{example}
 Consider the examples:
 \begin{align*}
 \mathrm{Pos}=\{f(4), f(6), f(8)\}\\
 \mathrm{Neg}=\{f(2), f(5), f(11)\}
 \end{align*}
and the partial program $H$:\\
\begin{center}
\emph{$H$: f(A) $\leftarrow$ geq(A,B), @numerical(B).}
\end{center}
There is no value for $B$ such that, once substituted in $H$, the resulting hypothesis is consistent (i.e. covers no negative examples) and covers at least one positive example. Therefore, we can prune its specialisations which do not have additional numerical literals. In particular, we prune the following partial programs $H_1$ and $H_2$:
\begin{center}
\begin{tabular}{l}
\emph{$H_1$: f(A) $\leftarrow$ geq(A,B), @numerical(B), odd(A).}\\
\emph{$H_2$: f(A) $\leftarrow$ geq(A,B), @numerical(B), div3(A).}
\end{tabular}
\end{center}
We provide a counter-example to explain why we cannot prune specialisation with additional numerical literals. Consider the following partial program $H_3$:
\begin{center}
\emph{$H_3$: f(A) $\leftarrow$ geq(A,B), @numerical(B), leq(A,C), @numerical(C).}
\end{center}
If applying the substitution $B=6$ and $C=10$, the resulting hypothesis covers the last positive example and no negative example.
\end{example}

\section{Implementation} \label{implementation}
\subsection{Numerical literals}
We provide \name{} with four built-in numerical literals shown in Figure 2 (\emph{geq}, \emph{leq}, \emph{add} and \emph{mult}). The user currently cannot define new numerical literals, since it would be cumbersome to specify their usage and definitions in SMT format. Moreover, as shown in Figure 3, the four built-in literals provided to \name{} already cover usual linear arithmetic fragments. We could consider adding more built-in literals or allowing the user to add custom ones if there is a need for some applications.

We set a bound over the maximum number of numerical literal per clause. This bound is a user parameter with a default value of 2. This value has been chosen as it supports most problems encountered.

The user can specify domains for the variables of numerical literals as part of the bias file, if these domains are known. This information is optional but can improve learning performance as it guides the search. The user also can specify the type of variables (currently supported types are \emph{real} or \emph{int}). There is a default value of \emph{real}. 
\section{Experiments}
\label{exp}
\subsection{Domain and Tasks.}
We describe the characteristics of the domains and tasks used in our experiments in Tables \ref{tab:domains} and \ref{tab:tasks}. Numerical values in target hypotheses are randomised for each run. Figure \ref{fig:sols} shows example solutions for each of the tasks. Some of these tasks require identifying non-numerical constant symbols. All systems tested support learning programs with constant symbols. For instance, solving \emph{zendo3} involves identifying the particular color \emph{blue}. Other tasks require learning recursive programs, which \name{} and \magicpopper{} fully support but \ale{} only partially supports.

\begin{table*}
\footnotesize
\centering
\begin{tabular}{l|ccc@{}}
\textbf{Domain} & \textbf{\# pos examples} & \textbf{\# neg examples} & \textbf{\# relations in bk}\\ \midrule
\emph{geometry} & 30 & 30 & 4\\ \midrule
\emph{zendo} & 30 & 30 & 11 \\ \midrule
\emph{pharma} & 30 & 30 & 9 \\\midrule
\emph{program synthesis} & 10 & 10 & 14\\
\end{tabular}
\caption{Domains description.}
\label{tab:domains}
\end{table*}

\begin{table*}
\footnotesize
\centering
\begin{tabular}{l|ccc@{}}
\textbf{Task} & \textbf{\# clauses} & \textbf{\# literals} & \textbf{\# numerical predicates}\\ \midrule
\emph{interval} & 1 & 3 & 2 \\
\emph{half plane} & 1 & 4 & 3 \\ \midrule
\emph{zendo1} & 1 & 5 & 1 \\
\emph{zendo2} & 2 & 10 & 4 \\
\emph{zendo3} & 2 & 11 & 4 \\
\emph{zendo4} & 3 & 15 & 5 \\
\midrule
\emph{pharma1} & 1 & 5 & 1 \\
\emph{pharma2} & 2 & 10 & 2\\
\emph{pharma3} & 2 & 12 & 3\\
\emph{pharma4} & 3 & 18 & 6\\
\midrule
\emph{member$\_$between} & 2 & 7 & 2\\
\emph{last$\_$leq} & 2 & 8 & 1\\
\emph{next$\_$geq} & 2 & 8 & 1\\
\end{tabular}
\caption{Learning tasks description.}
\label{tab:tasks}
\end{table*}

\begin{figure*}
\centering
\footnotesize
\begin{lstlisting}[caption=interval\label{interval}]
interval(A):- leq(9816.37), geq(D,4827.12).
\end{lstlisting}

\begin{lstlisting}[caption=halfplane\label{halfplane}]
halfplane(A,B) :- mult(A,3,D), add(B,D,E), leq(E,6).
\end{lstlisting}

\begin{lstlisting}[caption=zendo1\label{zendo1}]
zendo1(A) :- piece(A,B), contact(B,C), size(C,D), geq(E,7.32).
\end{lstlisting}

\begin{lstlisting}[caption=zendo2\label{zendo2}]
zendo2(A) :- piece(A,B), position(B,C,D), add(C,D,E), leq(E,6.92).
zendo2(A) :- piece(A,B), rotation(B,D), leq(D,4.12), geq(D, 3.23).
\end{lstlisting}

\begin{lstlisting}[caption=zendo3\label{zendo3}]
zendo3(A):- piece(A,B), position(B,C,D), leq(C,4.82), leq(D,2.61).
zendo3(A):- piece(A,B), color(B,blue), size(B,C), leq(C,6.39), geq(C,2.96).\end{lstlisting}

\begin{lstlisting}[caption=zendo4\label{zendo4}]
zendo4(A):- piece(A,B), size(B,C), geq(C,1.23), leq(C,4.66).
zendo4(A):- piece(A,B), position(B,X,Y), leq(X,3.45), leq(Y,6.87).
zendo4(A):- piece(A,B), contact(B,C), rotation(C,D), leq(D,1.18).
\end{lstlisting}

\begin{lstlisting}[caption=pharma1\label{pharma1}]
pharma1(A):- zincsite(A,E),hacc(A,D),dist(A,E,D,B),leq(B,4.98).
\end{lstlisting}

\begin{lstlisting}[caption=pharma2\label{pharma2}]
pharma2(A):- zincsite(A,B),hacc(A,C),dist(A,B,C,D),leq(B,2.94).
pharma2(A):- hacc(A,B),hacc(A,C),dist(A,B,C,D),geq(B,1.92).
\end{lstlisting}

\begin{lstlisting}[caption=pharma3\label{pharma3}]
pharma3(A):- zincsite(A,B), hacc(A,C), dist(A,B,C,D), leq(D,3.58), geq(D,1.78).
pharma3(A):- hacc(A,B), hacc(A,C), bond(A,B,C,du), dist(A,B,C,D), leq(D,2.78).
\end{lstlisting}

\begin{lstlisting}[caption=pharma4\label{pharma4}]
pharma4(A):- zincsite(A, B), hacc(A, C), dist(A, B, C, D), leq(D,4.18),geq(D,2.22).
pharma4(A):- hacc(A, C), hacc(A, E), dist(A, B, C, D), geq(D,1.23), leq(D,3.41).
pharma4(A):- zincsite(A, C), zincsite(A, B), bond(B,C,du), dist(A, B, C, D), leq(D,1.23).
\end{lstlisting}

\begin{lstlisting}[caption=member$\_$between\label{member_between}]
f(A):- head(A,D),leq(D,53),geq(D,45).
f(A):- tail(A,B),f(B).
\end{lstlisting}

\begin{lstlisting}[caption=last$\_$leq\label{last_leq}]
f(A):- tail(A,C),empty(C),head(A,B),leq(B,14).
f(A):- tail(A,B),f(B).
\end{lstlisting}

\begin{lstlisting}[caption=next$\_$geq\label{next_geq}]
f(A):- head(A,19),tail(A,E),head(E,C),geq(C,27).
f(A):- tail(A,B),f(B).
\end{lstlisting}

\caption{Example solutions.}
\label{fig:sols}
\end{figure*}

\subsection{Systems}
\paragraph{\name{} and \popper.}
We use for both systems the version \popperplus{} of \popper{} which learns non-separable programs and combines them \cite{popper+}. We allow \name{} and \magicpopper{} to use numerical predicates shown in Figure 2 (\emph{geq}, \emph{leq}, \emph{mult}, \emph{add}). 
We set for both systems the maximum number of numerical predicates in a clause to 2, apart from \emph{halfplane} for which we set it to 3. \magicpopper{} cannot learn constant symbols in non-deterministic predicates with a large or infinite number of answer substitutions, such as \emph{greater than}. Therefore, we disallow output constant symbols in numerical predicates for \magicpopper. However, we allow \magicpopper{} to learn constant symbols for other variables of type real or integer. 
For fairness in the experimental comparison, we disallow parallelisation in the solver during the \emph{numerical search} stage for \name{}. However, this setting can easily be changed to improve performance on real-world applications.

\paragraph{\ale.} We allow lazy evaluation for any of the numerical predicates. Since \ale{} does not automatically account for non-determinate predicates, we tailor the background definitions accordingly.

\subsection{Experimental Set-up}
Figures \ref{fig:biasnumsynth}, \ref{fig:biasmagic} and \ref{fig:biasale} show the experimental files for the task \emph{zendo1} for the systems evaluated. \magicpopper{} can learn constant symbols. Variables which are allowed to be constant symbols are flagged with the predicate \emph{magic$\_$value$\_$type}, which means that any variable of the type specified can be a constant symbol. \name{} additionally uses \emph{numerical$\_$predicates}. This predicate indicates which built-in numerical predicates are allowed in the learning task. These user can define bounds for the numerical variables in numerical predicates through the predicate \emph{bounds}, although this information is optional.

The experimental files also include magic values declarations. Magic variables can be variables of any types and can appear in any background predicate. For instance, magic variables can be used to find constant symbols which are not numeric, such as string or lists, from user defined background relations. By contrast, numerical predicates are restricted to built-in numerical predicates, and they must have type \emph{real} or \emph{int}. However, the use of numerical predicates allows more complex and elaborated numerical reasoning from multiple examples, but this numerical reasoning comes at the expense of more  complexity. 



\begin{figure}
\centering
\footnotesize
\begin{lstlisting}
max_vars(6).
max_body(5).

head_pred(zendo,1).
body_pred(piece,2).
body_pred(color,2).
body_pred(size,2).
body_pred(position,3).
body_pred(rotation,2).
body_pred(orientation,2).
body_pred(contact,2).

type(zendo,(state,)).
type(piece,(state,piece)).
type(color,(piece,color)).
type(size,(piece,real)).
type(position,(piece,real,real)).
type(rotation,(piece,real)).
type(orientation,(piece,orientation)).
type(contact,(piece,piece)).

direction(zendo,(in,)).
direction(piece,(in,out)).
direction(color,(in,out)).
direction(size,(in,out)).
direction(position,(in,out,out)).
direction(rotation,(in,out)).
direction(orientation,(in,out)).
direction(contact,(in,out)).

magic_value_type(color).
magic_value_type(orientation).

numerical_pred(geq,2).
numerical_pred(leq,2).

type(geq,(real,real)).
type(leq,(real,real)).

direction(geq,(in, out)).
direction(leq,(in, out)).

numerical_pred(add,3).
type(add,(real, real, real)).
direction(add,(in,in,out)).

numerical_pred(mult,3).
type(mult,(real, real, real)).
direction(mult,(in,out,in)).

bounds(geq,1,(-10,10)).
bounds(leq,1,(-10,10)).
bounds(mult,1,(-10,10)).
bounds(add,1,(-10,10)).
\end{lstlisting}
\caption{Example of experimental set-up for \name{} for the learning task \emph{zendo1}.}
\label{fig:biasnumsynth}
\end{figure}

\begin{figure}
\begin{lstlisting}
max_vars(6).
max_body(5).

head_pred(zendo,1).
body_pred(piece,2).
body_pred(color,2).
body_pred(size,2).
body_pred(position,3).
body_pred(rotation,2).
body_pred(orientation,2).
body_pred(contact,2).

type(zendo,(state,)).
type(piece,(state,piece)).
type(color,(piece,color)).
type(size,(piece,real)).
type(position,(piece,real,real)).
type(rotation,(piece,real)).
type(orientation,(piece,orientation)).
type(contact,(piece,piece)).

direction(zendo,(in,)).
direction(piece,(in,out)).
direction(color,(in,out)).
direction(size,(in,out)).
direction(position,(in,out,out)).
direction(rotation,(in,out)).
direction(orientation,(in,out)).
direction(contact,(in,out)).

magic_value_type(color).
magic_value_type(orientation).
magic_value_type(real).

body_pred(my_geq,2).
body_pred(my_leq,2).
body_pred(my_add,3).
body_pred(my_mult,3).

type(my_geq,(real,real)).
type(my_leq,(real,real)).
type(my_add,(real,real,real)).
type(my_mult,(real,real,real)).

direction(my_geq,(in, in)).
direction(my_leq,(in, in)).
direction(my_add,(in, in, out)).
direction(my_mult,(in, in, out)).

num_p(my_geq,2).
num_p(my_leq,2).
num_p(my_add,3).
num_p(my_mult,3).

incompatible(my_geq,my_geq).
incompatible(my_leq,my_leq).
incompatible(eq,my_geq).
incompatible(eq,leq).

:- clause(C), #count{P,Vars : body_literal(C,P,A,Vars), num_p(P,A)} > 2.
\end{lstlisting}
\caption{Example of experimental set-up for \magicpopper{} for the learning task \emph{zendo1}.}
\label{fig:biasmagic}
\end{figure}

\begin{figure}
\begin{lstlisting}
:- aleph_set(verbosity, 1).
:- aleph_set(interactive, false).
:- aleph_set(i,4).
:- aleph_set(clauselength,6).
:- aleph_set(nodes,10000).

:- modeh(*,zendo(+state)).
:- modeb(*,piece(+state,-piece)).
:- modeb(*,color(+piece,#color)).
:- modeb(*,size(+piece,-real)).
:- modeb(*,position(+piece,-real,-real)).
:- modeb(*,rotation(+piece,-real)).
:- modeb(*,orientation(+piece,#orientation)).
:- modeb(*,contact(+piece,-piece)).
:- modeb(*,my_geq(+real,#real)).
:- modeb(*,my_leq(+real,#real)).
:- modeb(*,my_add(+real,+real,-real)).
:- modeb(*,my_mult(+real,#real,-real)).

:- determination(zendo/1,piece/2).
:- determination(zendo/1,color/2).
:- determination(zendo/1,size/2).
:- determination(zendo/1,position/3).
:- determination(zendo/1,rotation/2).
:- determination(zendo/1,orientation/2).
:- determination(zendo/1,contact/2).
:- determination(zendo/1,my_geq/2).
:- determination(zendo/1,my_leq/2).
:- determination(zendo/1,my_add/3).
:- determination(zendo/1,my_mult/3).

:- lazy_evaluate(my_geq/2).
:- lazy_evaluate(my_leq/2).
\end{lstlisting}
\caption{Example of experimental set-up for \ale{} for the learning task \emph{zendo1}.}
\label{fig:biasale}
\end{figure}

\subsection{Results: Scalability}
For successive values of $n$, we sample $n$ positive and $n$ negative examples. Figures \ref{fig:acc} and \ref{fig:pharm_acc} represent the predictive accuracies versus the number of examples for \emph{pharma2}.
\definecolor{mygreen}{rgb}{0.1,0.6,0.1}
\pgfplotstableread[col sep=comma]{./zendoexamples/26_7_zendo_1__aleph.csv}\resaleph
\pgfplotstableread[col sep=comma]{./figures/new_combine/Zendo/magicpopper.csv}\resmagicpopper
\pgfplotstableread[col sep=comma]{./figures/new_combine/Zendo/numericalpopper.csv}\resnumericalpopper

\pgfplotstableread[col sep=comma]{./figures/new_combine/pharma/numericalpopper.csv}\resnumericalpopperpharma
\pgfplotstableread[col sep=comma]{./figures/new_combine/pharma/magicpopper.csv}\resmagicpopperpharma
\pgfplotstableread[col sep=comma]{./pharmaexamples/pharma2__aleph.csv}\resalephpharma


\begin{figure*}
  \begin{minipage}{0.1\textwidth}
\begin{tikzpicture}
\begin{customlegend}[legend columns=5,legend style={nodes={scale=1, transform shape},align=left,column sep=0ex},
        legend entries={\name, \ale, \magicpopper}]
        \addlegendimage{red,mark=diamond*}
        \addlegendimage{blue,mark=square*}
        \addlegendimage{mygreen,mark=triangle*}
\end{customlegend}
\end{tikzpicture}
\end{minipage}\hfill\\
  \begin{minipage}{0.4\textwidth}
\resizebox{\columnwidth}{!}{
\begin{tikzpicture}
\begin{axis}[
  legend style={at={(0.5,0.5)},anchor=west},
  legend style={font=\normalsize},
  xtick={0,100,200,300,400,500},
  xlabel=Number of examples,
  xlabel style={font=\Large},
  ylabel style={font=\Large},
  tick label style={font=\Large},  
  ylabel=Accuracy (\%),
  ]
  \addplot+[blue,mark=square*,
                error bars/.cd,
                y dir=both,
                error mark,
                y explicit]table[x=xs,y=acc_av,
  y error=acc_std] {\resaleph};
    \addplot[red,mark=diamond*,
                error bars/.cd,
                y dir=both,
                error mark,
                y explicit]table[x=xs,y=acc_av,
    y error=acc_std] {\resnumericalpopper};
    \addplot[mygreen,mark=triangle*,
                error bars/.cd,
                y dir=both,
                error mark,
                y explicit]table[x=xs ,y=acc_av,y error=acc_std] {\resmagicpopper};
\end{axis}
\end{tikzpicture}}
\caption{Accuracy versus the number of examples for \emph{zendo1}.}
\label{fig:acc}
  \end{minipage}\hfill
  \begin{minipage}{0.4\textwidth}
\resizebox{\columnwidth}{!}{
\begin{tikzpicture}
\begin{axis}[
  legend style={at={(0.5,0.5)},anchor=west},
  legend style={font=\normalsize},
  xtick={0,100,200,300,400,500},
  xlabel=Number of examples,
  xlabel style={font=\Large},
  ylabel style={font=\Large},
  tick label style={font=\Large},  
  ylabel=Accuracy (\%),
  ]
    \addplot[red,mark=diamond*,
                error bars/.cd,
                y dir=both,
                error mark,
                y explicit]table[x=xs,y=acc_av,
    y error=acc_std] {\resnumericalpopperpharma};
    \addplot[mygreen,mark=triangle*,
                error bars/.cd,
                y dir=both,
                error mark,
                y explicit]table[x=xs,y=acc_av,
    y error=acc_std] {\resmagicpopperpharma};
    \addplot[blue,mark=square*,
                error bars/.cd,
                y dir=both,
                error mark,
                y explicit]table[x=xs,y=acc_av,
    y error=acc_std] {\resalephpharma};
\end{axis}
\end{tikzpicture}}
\caption{Accuracy versus the number of examples for \emph{pharma2}.}
\label{fig:pharm_acc}
  \end{minipage}
\end{figure*}


\subsection{Results: Learning Time}
Figure \ref{proportion} represents the proportion of the learning time spent in the different stages. It shows the learning time is dominated by the numerical stage: about 40\% on average over tasks of the learning time is used to build and solve SMT formula, and about 6.5\% is used to find substitutions for the related variables with Prolog.

\pgfplotstableread[col sep=comma,header=true]{
stages,1,2,3,4,5
interval,47.9623393,10.2207694,1.62734284,1.19657167,38.9929768
halfplane,75.6208319,6.06841385,1.84621776,9.56507831,6.89945818
zendo1,54.73022817,3.61407717,6.4437201,31.5218948,3.69007974
zendo2,44.93339374,2.31203429,7.23109319,42.1067995,3.41667928
zendo3,39.42860857,5.60064573,17.5605139,34.6676627,2.74256903
zendo4,42.27688618,6.17383869,15.9259426,32.9178792,2.70545338
pharma1,26.68816362,7.89573589,4.03502297,30.00774,31.3733376
pharma2,25.88394646,9.90349379,13.3959605,28.8944157,21.9221836
pharma3,24.45073368,3.53134137,8.91401473,59.5863289,3.51758137
pharma4,18.83914918,3.51423639,8.7316383,65.4264795,3.4884966
member$\_$between,47.79124698,8.94230953,13.130869,16.378686,13.7568886
last$\_$leq,43.92335924,9.60160347,16.9897479,22.6400366,6.84525286
next$\_$geq,64.04517566,10.5980003,7.22199596,4.29849546,13.8363326

}\data

 \pgfplotsset{compat=1.14}
 \definecolor{stage1}{rgb}{0.54, 0.81, 0.94}
 \definecolor{stage2}{rgb}{1.0, 0.72, 0.77}
\definecolor{stage3}{rgb}{0.67, 0.88, 0.69}
\definecolor{stage4}{rgb}{1.0, 0.99, 0.82}
\definecolor{stage5}{rgb}{0.76, 0.69, 0.57}

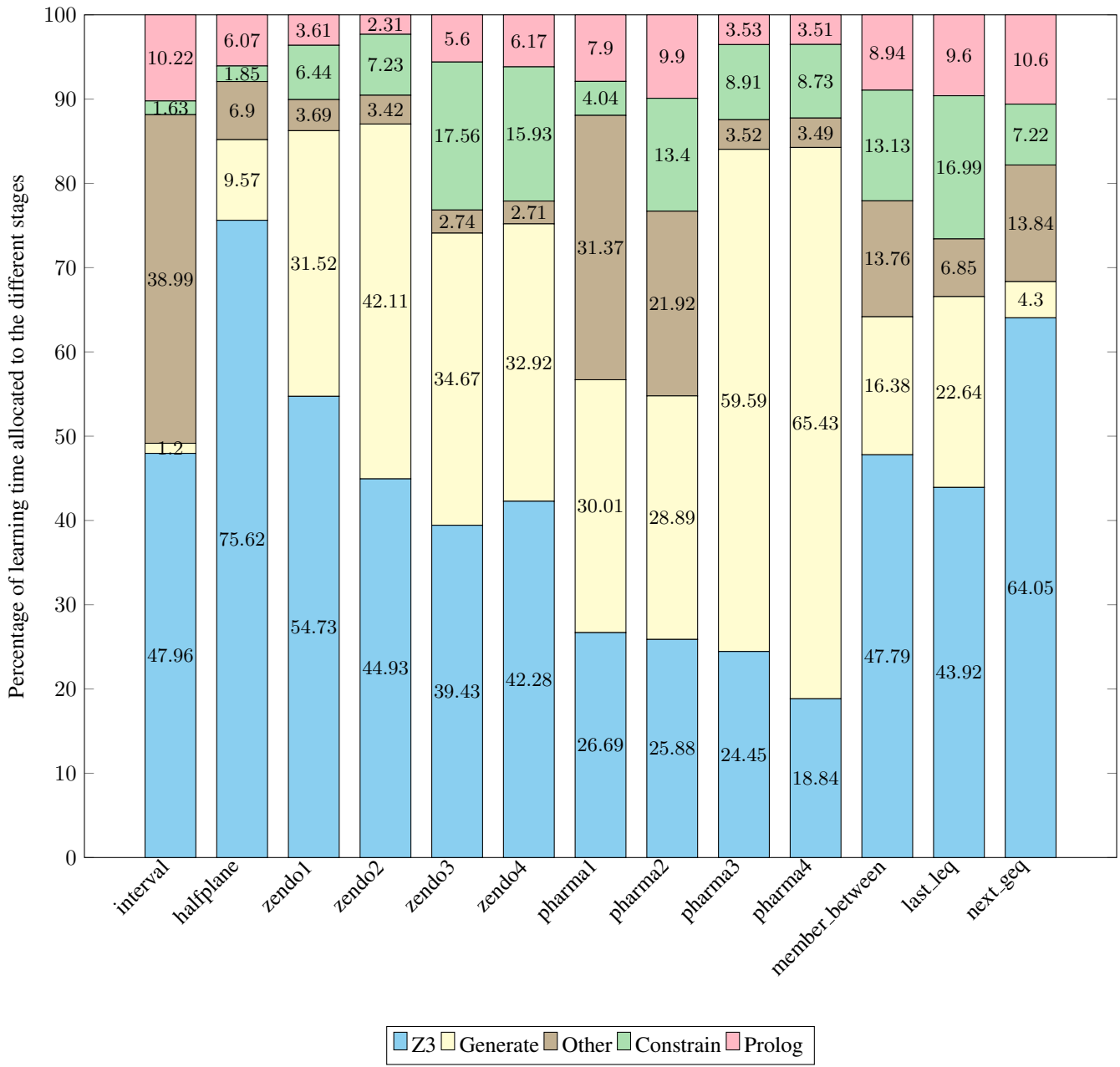
\begin{figure*}
\begin{tikzpicture}
\begin{axis}[
    ybar stacked,
	bar width=23pt,
	height=15cm,
	width=18cm,
	nodes near coords,
	nodes near coords style={font=\footnotesize},
	ymin=0,ymax=100,
    legend style={at={(0.5,-0.20)},
      anchor=north,legend columns=-1},
    ylabel={Percentage of learning time allocated to the different stages},
    symbolic x coords={interval,halfplane,zendo1,zendo2,zendo3,zendo4,pharma1,pharma2,pharma3,pharma4,member$\_$between,last$\_$leq,next$\_$geq},
    xtick=data,
    xticklabels from table={\data}{stages},
    x tick label style={rotate=45,anchor=east},
    ]

\addplot [fill=stage1] table [y=1]  {\data};
            addlegendentry{1};
\addplot [fill=stage4] table [y=4]  {\data};
            addlegendentry{1};
\addplot [fill=stage5] table [y=5]  {\data};
            addlegendentry{1};
\addplot [fill=stage3] table [y=3]  {\data};
            addlegendentry{1};
\addplot [fill=stage2] table [y=2]  {\data};
            addlegendentry{1};
\legend{\strut Z3, \strut Generate, \strut Other, \strut Constrain, \strut Prolog}
\end{axis}
\end{tikzpicture}
\caption{Proportion of learning time spent in each of the different stages. \emph{Z3} refers to building and solving the SMT formula. \emph{Generate} refers to the program generation stage. \emph{Constrain} refers to the generation of constraints to prune the hypothesis space. \emph{Prolog} refers to finding substitutions for the related variables. \emph{Other} refers to any other operation.}
\label{proportion}
\end{figure*}

\bibliography{manuscript}

\begin{thebibliography}{37}
\providecommand{\natexlab}[1]{#1}

\bibitem[{Albarghouthi et~al.(2017)Albarghouthi, Koutris, Naik, and
  Smith}]{albarghouthi2017}
Albarghouthi, A.; Koutris, P.; Naik, M.; and Smith, C. 2017.
\newblock Constraint-based synthesis of Datalog programs.
\newblock In \emph{International Conference on Principles and Practice of
  Constraint Programming}, 689--706. Springer.

\bibitem[{Anthony and Frisch(1997)}]{anthony1997}
Anthony, S.; and Frisch, A.~M. 1997.
\newblock Generating numerical literals during refinement.
\newblock In \emph{International Conference on Inductive Logic Programming},
  61--76.

\bibitem[{Blockeel and De~Raedt(1997)}]{blockeel1997}
Blockeel, H.; and De~Raedt, L. 1997.
\newblock Lookahead and discretization in ILP.
\newblock In \emph{Inductive Logic Programming}, 77--84.

\bibitem[{Blockeel and De~Raedt(1998)}]{blockeel1998}
Blockeel, H.; and De~Raedt, L. 1998.
\newblock Top-down induction of first-order logical decision trees.
\newblock \emph{Artificial Intelligence}, 101(1-2): 285--297.

\bibitem[{Corapi, Russo, and Lupu(2011)}]{corapi2011}
Corapi, D.; Russo, A.; and Lupu, E. 2011.
\newblock {I}nductive {L}ogic {P}rogramming in {A}nswer {S}et {P}rogramming.
\newblock In \emph{Inductive Logic Programming}, 91--97.

\bibitem[{Cropper(2022)}]{popper+}
Cropper, A. 2022.
\newblock Learning programs by combining programs.

\bibitem[{Cropper et~al.(2022)Cropper, Dumancic, Evans, and Muggleton}]{ilp30}
Cropper, A.; Dumancic, S.; Evans, R.; and Muggleton, S.~H. 2022.
\newblock Inductive logic programming at 30.
\newblock \emph{Mach. Learn.}, 111(1): 147--172.

\bibitem[{Cropper and Morel(2021)}]{cropper2020b}
Cropper, A.; and Morel, R. 2021.
\newblock Learning programs by learning from failures.
\newblock \emph{Mach. Learn.}, 1--56.

\bibitem[{De~Moura and Bj{\o}rner(2011)}]{smt}
De~Moura, L.; and Bj{\o}rner, N. 2011.
\newblock Satisfiability modulo theories: introduction and applications.
\newblock \emph{CACM}, 54(9): 69--77.

\bibitem[{Dietterich, Lathrop, and Lozano-P{\'e}rez(1997)}]{dietterich1997}
Dietterich, T.~G.; Lathrop, R.~H.; and Lozano-P{\'e}rez, T. 1997.
\newblock Solving the multiple instance problem with axis-parallel rectangles.
\newblock \emph{Artificial intelligence}, 89(1-2): 31--71.

\bibitem[{Ellis et~al.(2018)Ellis, Morales, Sabl\'{e}-Meyer, Solar-Lezama, and
  Tenenbaum}]{ellis2018}
Ellis, K.; Morales, L.; Sabl\'{e}-Meyer, M.; Solar-Lezama, A.; and Tenenbaum,
  J. 2018.
\newblock Learning Libraries of Subroutines for Neurally\textendash Guided
  Bayesian Program Induction.
\newblock In \emph{Advances in Neural Information Processing Systems},
  volume~31.

\bibitem[{Evans and Grefenstette(2018)}]{evans2018}
Evans, R.; and Grefenstette, E. 2018.
\newblock Learning explanatory rules from noisy data.
\newblock \emph{JAIR}, 61: 1--64.

\bibitem[{Evans et~al.(2021)Evans, Hern\'{a}ndez-Orallo, Welbl, Kohli, and
  Sergot}]{evans2021}
Evans, R.; Hern\'{a}ndez-Orallo, J.; Welbl, J.; Kohli, P.; and Sergot, M. 2021.
\newblock Making sense of sensory input.
\newblock \emph{Artificial Intelligence}, 293: 103438.

\bibitem[{Finn et~al.(1998)Finn, Muggleton, Page, and Srinivasan}]{finn1998}
Finn, P.; Muggleton, S.; Page, D.; and Srinivasan, A. 1998.
\newblock Pharmacophore Discovery Using the Inductive Logic Programming System
  PROGOL.
\newblock \emph{Mach. Learn.}, 30(2): 241--270.

\bibitem[{Gebser et~al.(2014)Gebser, Kaminski, Kaufmann, and
  Schaub}]{gebser2014}
Gebser, M.; Kaminski, R.; Kaufmann, B.; and Schaub, T. 2014.
\newblock Clingo= {ASP}+ control: Preliminary report.
\newblock \emph{arXiv preprint arXiv:1405.3694}.

\bibitem[{Gulwani et~al.(2011)Gulwani, Jha, Tiwari, and
  Venkatesan}]{gulwani2011b}
Gulwani, S.; Jha, S.; Tiwari, A.; and Venkatesan, R. 2011.
\newblock Synthesis of loop-free programs.
\newblock \emph{ACM SIGPLAN Notices}, 46(6): 62--73.

\bibitem[{Hocquette and Cropper(2022)}]{hocquette2022}
Hocquette, C.; and Cropper, A. 2022.
\newblock Learning programs with magic values.
\newblock \emph{arXiv preprint arXiv:2208.03238}.

\bibitem[{Jha et~al.(2010)Jha, Gulwani, Seshia, and Tiwari}]{jha2010}
Jha, S.; Gulwani, S.; Seshia, S.~A.; and Tiwari, A. 2010.
\newblock Oracle-guided component-based program synthesis.
\newblock In \emph{2010 ACM/IEEE 32nd International Conference on Software
  Engineering}, volume~1, 215--224. IEEE.

\bibitem[{Kaminski, Eiter, and Inoue(2018)}]{kaminski2018}
Kaminski, T.; Eiter, T.; and Inoue, K. 2018.
\newblock Exploiting Answer Set Programming with External Sources for
  Meta-Interpretive Learning.
\newblock \emph{Theory and Practice of Logic Programming}, 18(3-4): 571--588.

\bibitem[{Karali{\v{c}} and Bratko(1997)}]{karalivc1997}
Karali{\v{c}}, A.; and Bratko, I. 1997.
\newblock First order regression.
\newblock \emph{Mach. Learn.}, 26(2): 147--176.

\bibitem[{Lloyd(2012)}]{lloyd:book}
Lloyd, J.~W. 2012.
\newblock \emph{Foundations of logic programming}.
\newblock Springer Science \& Business Media.

\bibitem[{Moura and Bj{\o}rner(2008)}]{moura2008}
Moura, L.~d.; and Bj{\o}rner, N. 2008.
\newblock Z3: An efficient SMT solver.
\newblock In \emph{International conference on Tools and Algorithms for the
  Construction and Analysis of Systems}, 337--340. Springer.

\bibitem[{Muggleton(1991)}]{muggleton1991}
Muggleton, S.~H. 1991.
\newblock Inductive logic programming.
\newblock \emph{New Generation Computing}, 8(4): 295--318.

\bibitem[{Muggleton(1995)}]{muggleton1995}
Muggleton, S.~H. 1995.
\newblock Inverse {E}ntailment and {P}rogol.
\newblock \emph{New Generation Comput.}, 13(3{\&}4): 245--286.

\bibitem[{Muggleton and De~Raedt(1994)}]{muggleton1994}
Muggleton, S.~H.; and De~Raedt, L. 1994.
\newblock Inductive Logic Programming: Theory and Methods.
\newblock \emph{The Journal of Logic Programming}, 19-20: 629--679.

\bibitem[{Muggleton, Lin, and Tamaddoni-Nezhad(2015)}]{muggleton2015}
Muggleton, S.~H.; Lin, D.; and Tamaddoni-Nezhad, A. 2015.
\newblock Meta-Interpretive Learning of Higher-Order Dyadic Datalog: Predicate
  Invention revisited.
\newblock \emph{Mach. Learn.}, 100(1): 49--73.

\bibitem[{Purga{\l}, Cerna, and Kaliszyk(2022)}]{hopper}
Purga{\l}, S.~J.; Cerna, D.~M.; and Kaliszyk, C. 2022.
\newblock Learning Higher-Order Logic Programs From Failures.
\newblock In \emph{{IJCAI} 2022}, 2726--2733.

\bibitem[{Raghothaman et~al.(2019)Raghothaman, Mendelson, Zhao, Naik, and
  Scholz}]{raghothaman2019}
Raghothaman, M.; Mendelson, J.; Zhao, D.; Naik, M.; and Scholz, B. 2019.
\newblock Provenance-guided synthesis of Datalog programs.
\newblock \emph{Proceedings of the ACM on Programming Languages}, 4(POPL):
  1--27.

\bibitem[{Reynolds et~al.(2015)Reynolds, Deters, Kuncak, Tinelli, and
  Barrett}]{reynolds2015}
Reynolds, A.; Deters, M.; Kuncak, V.; Tinelli, C.; and Barrett, C. 2015.
\newblock Counterexample-guided quantifier instantiation for synthesis in SMT.
\newblock In \emph{International Conference on Computer Aided Verification},
  198--216. Springer.

\bibitem[{Sch{\"u}ller and Benz(2018)}]{schuller2018}
Sch{\"u}ller, P.; and Benz, M. 2018.
\newblock Best-effort inductive logic programming via fine-grained cost-based
  hypothesis generation.
\newblock \emph{Mach. Learn.}, 107(7): 1141--1169.

\bibitem[{Sebag and Rouveirol(1996)}]{sebag1996}
Sebag, M.; and Rouveirol, C. 1996.
\newblock Constraint inductive logic programming.

\bibitem[{Shi et~al.(2022)Shi, Dai, Ellis, and Sutton}]{crossbeam}
Shi, K.; Dai, H.; Ellis, K.; and Sutton, C. 2022.
\newblock CrossBeam: Learning to search in bottom-up program synthesis.
\newblock \emph{arXiv preprint arXiv:2203.10452}.

\bibitem[{Solar-Lezama(2009)}]{solar2009}
Solar-Lezama, A. 2009.
\newblock The sketching approach to program synthesis.
\newblock In \emph{Asian Symposium on Programming Languages and Systems},
  4--13. Springer.

\bibitem[{Srinivasan(2001)}]{srinivasan2001}
Srinivasan, A. 2001.
\newblock The {ALEPH} manual.
\newblock \emph{Machine Learning at the Computing Laboratory}.

\bibitem[{Srinivasan and Camacho(1999)}]{srinivasan1999}
Srinivasan, A.; and Camacho, R. 1999.
\newblock Numerical reasoning with an ILP system capable of lazy evaluation and
  customised search.
\newblock \emph{The Journal of Logic Programming}, 40(2): 185--213.

\bibitem[{Srinivasan et~al.(2006)Srinivasan, Page, Camacho, and
  King}]{srinivasan2006}
Srinivasan, A.; Page, D.; Camacho, R.; and King, R. 2006.
\newblock Quantitative pharmacophore models with inductive logic programming.
\newblock \emph{Mach. Learn.}, 64(1): 65--90.

\bibitem[{Wahlig(2022)}]{wahlig2022}
Wahlig, J. 2022.
\newblock Learning Logic Programs From Noisy Failures.
\newblock \emph{CoRR}, abs/2201.03702.

\end{thebibliography}

\end{document}